\documentclass[]{style}

\usepackage{graphicx}
\usepackage{svg}
\usepackage{float}
\usepackage{wrapfig}
\usepackage{subcaption}
\usepackage{caption}
\usepackage{adjustbox}
\usepackage{tikz}
\usetikzlibrary{positioning, fit}

\usepackage{array}
\usepackage{booktabs}
\usepackage{tabularx}
\usepackage{multirow}
\usepackage{makecell}
\usepackage{colortbl}

\usepackage[table]{xcolor}
\definecolor{headergray}{gray}{0.90}
\definecolor{rowgray}{gray}{0.95}
\definecolor{agentColor}{RGB}{0, 51, 153}
\definecolor{assistColor}{RGB}{128, 0, 128}
\definecolor{failRed}{RGB}{200, 0, 0}
\definecolor{jsonKey}{RGB}{0, 100, 0}
\definecolor{strColor}{RGB}{42, 0, 255}
\definecolor{codebg}{RGB}{245,246,247}
\definecolor{codeframe}{RGB}{200,200,200}

\usepackage{amsmath}
\usepackage{amssymb}
\usepackage{mathtools}
\usepackage{amsthm}

\usepackage{fontawesome5}
\usepackage{inconsolata}
\usepackage{microtype}

\usepackage{enumitem}
\usepackage{ragged2e}
\usepackage{comment}

\usepackage{listings}
\lstdefinestyle{chattemplate}{
  basicstyle=\ttfamily\small,
  breaklines=true,
  breakatwhitespace=true,
  showstringspaces=false,
  columns=fullflexible,
  frame=none,
  xleftmargin=0pt,
  aboveskip=0pt,
  belowskip=0pt
}

\usepackage[most]{tcolorbox}
\tcbuselibrary{listings,breakable,skins}

\usepackage[toc,page,header]{appendix}

\usepackage{hyperref}

\theoremstyle{plain}

\theoremstyle{definition}

\theoremstyle{remark}

\title{Source-Free MT Evaluation Is Not MT Evaluation}

\renewcommand{\authorlist}{%
  \authorfont\sffamily\bfseries
  Baban Gain$^{1}\dag$, Ramakrishna Appicharla$^{2}$, Asif Ekbal$^{1}$%
}

\affiliation[1]{Indian Institute of Technology Patna}
\affiliation[2]{Shiv Nadar University, Chennai}

\contribution[\dag]{Corresponding Author}

\abstract{
Reference-based metrics remain the standard choice in machine translation evaluation, partly because quality estimation methods often correlate less well with human judgments. As a result, source-free, reference-based evaluation has become the practical norm, even though it is unfaithful to the definition of translation adequacy and unfair to systems whose outputs preserve the source meaning while differing from the reference. This paper argues that adequacy must be judged with respect to the source. A reference is only one possible rendering of the source and may introduce bias, under-specification, or errors. We further argue that source-reference-hypothesis evaluation is fair only when the judge treats the reference as auxiliary evidence rather than as the primary standard. Otherwise, even source-aware evaluation can reduce adequacy to preference towards reference. We show the existing hybrid metrics are highly reliant on reference compared to source. Our argument is not that all automatic MT metrics fail to use the source. Rather, we argue that any evaluation protocol that removes the source, or allows the reference to dominate the source, is structurally incomplete for adequacy evaluation. However, existing MT papers generally prefer reference-based metrics and use QE metrics only when reference is unavailable. We therefore call for QE to be reframed as a primary approach to source-grounded adequacy evaluation, rather than as a fallback motivated by missing references. We further call for hybrid metrics whose designs explicitly prioritize source--hypothesis faithfulness while using references only as complementary evidence.
}
\checkdata[\faEnvelope~Contact]{gainbaban@gmail.com}

\begin{document}

\maketitle

\section{Introduction}

Machine translation (MT) evaluation asks whether a system output is a good
translation of a given source sentence. Reliable evaluation is essential for
comparing systems, selecting models, guiding development, and measuring
progress. Yet translation quality is difficult to assess automatically because
it encompasses several related but distinct properties, most prominently
fluency in the target language and adequacy with respect to the source. Human
evaluation can assess these properties directly, but it is costly,
time-consuming, and difficult to reproduce at scale. Automatic metrics have
therefore become the primary means of evaluating MT systems.

Automatic MT evaluation has historically been dominated by reference-based
metrics. Metrics such as BLEU and chrF compare a system hypothesis with one or
more human-written reference translations using lexical overlap
\citep{papineni-etal-2002-bleu,popovic-2015-chrf}. Later metrics, including
BERTScore, BLEURT, and COMET, use contextual representations, supervision from
human judgments, or both to capture similarities beyond exact lexical matches
\citep{Zhang*2020BERTScore:,sellam-etal-2020-bleurt,rei-etal-2020-comet}. Although these
metrics differ substantially in their architectures and training objectives,
they share a common premise: agreement with a human reference provides evidence
of translation quality.
A reference, however, represents only one possible translation of the source.
The same meaning can often be expressed through different lexical choices,
syntactic structures, levels of explicitness, or culturally appropriate
reformulations. A hypothesis may therefore preserve the source meaning while
differing considerably from the available reference. Conversely, similarity to
a reference does not by itself establish that the hypothesis faithfully
translates the source, particularly when the reference is ambiguous,
underspecified, noisy, or erroneous. Reference-based evaluation can therefore
conflate translation quality with similarity to one particular human rendering
\citep{freitag-etal-2020-bleu}.

Reference-free evaluation, commonly studied as machine translation quality
estimation (QE), removes the need for a reference and predicts translation
quality from the source and hypothesis
\citep{specia-etal-2013-quest,rei-etal-2022-cometkiwi}. QE is generally
motivated by the practical limitations of references: human translations are
costly to produce, unavailable in many deployment settings, and difficult to
obtain for new domains and low-resource languages. Reference-free evaluation is
accordingly presented primarily as an alternative for settings in which a
reference is unavailable, rather than as the conceptually appropriate setting
for determining whether a hypothesis preserves the meaning of its source. Even
when faithfulness is included as an evaluation dimension, the absence of a
reference is often treated as a limitation that the metric must overcome.

We argue that this framing reverses the underlying relationship. A hypothesis
should be evaluated against its source not merely because a reference may be
unavailable, but because translation adequacy is defined with respect to the
source. The source is therefore not a substitute for a missing reference. It is
the primary evidence required to determine whether the output is a faithful
translation. A reference can provide useful auxiliary evidence about fluency,
lexical realization, or one acceptable interpretation, but it cannot replace
the source as the authority on the meaning that must be preserved. From this
perspective, evaluation based only on the hypothesis and reference is not simply
reference-based MT evaluation. It is source-free evaluation of target-side
agreement.
This distinction also applies to hybrid metrics that receive the source,
hypothesis, and reference together. Providing the source as an input does not
guarantee that a metric grounds its judgment in the source. A model may formally
accept all three inputs while relying predominantly on hypothesis--reference
compatibility. In that case, the source has been added to the input without
becoming the basis of the adequacy judgment. The important question is therefore
not only which inputs a metric receives, but which input governs its decision
when the source and reference provide conflicting evidence.

In this paper, we argue for a source-grounded formulation of MT
evaluation in which adequacy is judged primarily from the source--hypothesis
relationship, while references, when available, serve only as auxiliary
evidence. We first distinguish adequacy from fluency and identify translation
phenomena that cannot be judged fairly without access to the source. We then
introduce a counterfactual replacement diagnostic to measure whether a hybrid
metric is more sensitive to the source or to the reference. Holding a known good
translation fixed, we separately replace its source and reference with
incompatible examples and compare the resulting score changes. If a metric
treats the source as the authority for adequacy, corrupting the source should be
at least as consequential as corrupting the reference.

We apply this diagnostic to prominent hybrid metrics, including COMET
\citep{rei-etal-2020-comet}, XCOMET-XXL
\citep{guerreiro-etal-2024-xcomet}, and MetricX-24-Hybrid-XL
\citep{juraska-etal-2024-metricx}. We find that their scores can be substantially
more sensitive to reference corruption than to source corruption. For COMET,
replacing the reference causes an average score change more than ten times as
large as replacing the source, and reference sensitivity exceeds source
sensitivity for 99.54\% of the analyzed examples. We further examine the same
problem in LLM-as-a-judge evaluation, contrast reference-dominated behavior with
more explicitly source-grounded evaluation protocols, and provide
recommendations for the design and reporting of MT evaluation. Our broader
claim is not that references are useless or that existing metrics completely
ignore the source. Rather, an evaluation protocol is structurally incomplete
for adequacy when it removes the source or permits the reference to dominate it.

\section{Related Work}
\label{sec:related-work}

\paragraph{\textbf{Reference bias in MT evaluation}}
Concerns about reference-based evaluation predate contemporary neural metrics. A human reference represents only one acceptable realization of the source, and evaluation against that realization can favor hypotheses that resemble it lexically or structurally. In human evaluation, \citet{fomicheva-specia-2016-reference} found that monolingual assessors can be influenced by the reference and argued that adequacy is better assessed against the original source. However, \citet{ma-etal-2017-investigation} reexamined this result and found no significant evidence of reference bias under their experimental conditions. Reference choice also affects automatic evaluation: \citet{freitag-etal-2020-bleu} showed that the construction and diversity of references can substantially influence metric reliability, particularly when evaluating high-quality systems. These studies establish that references can shape both human and automatic judgments, but they primarily investigate human evaluation protocols or the quality and coverage of the references themselves. Our concern is different: whether a metric that receives both the source and a reference allows the reference to govern its adequacy judgment.

\paragraph{\textbf{Source use in learned metrics}}
The development of learned metrics has made it possible to condition evaluation jointly on the source, hypothesis, and reference. Nevertheless, providing the source as an input does not ensure that it materially influences the resulting score. Using explanation methods for COMET and UniTE, \citet{rei-etal-2023-inside} found evidence that neural metrics may rely heavily on reference information while making limited use of the source. A related pattern has been observed in LLM-based evaluation. Through controlled comparisons of source-based, reference-based, and combined input settings, \citet{huang-etal-2024-lost} found that references consistently improved LLM evaluation performance, whereas source information was sometimes unhelpful or counterproductive. These findings motivate examining not only which inputs an evaluator receives, but also how its decisions depend on each input.

\paragraph{\textbf{Translation accuracy challenge sets.}}
The work most closely related to ours is ACES, introduced by
\citet{moghe-etal-2025-machine}. ACES evaluates whether MT metrics can distinguish good and incorrect translations across 68 accuracy phenomena and 146 language pairs. Several of its challenge sets are designed so that the reference is ambiguous or insufficient and the source is required to identify the better hypothesis. Evaluating 47 metrics, \citet{moghe-etal-2025-machine} find that reference-based metrics often fail to exploit source information and can be distracted by surface similarity to the reference. They consequently argue that the source is the primary textual evidence for translation accuracy and recommend that metrics focus more explicitly on it.

Our work builds directly on this conclusion but asks a different diagnostic question. ACES changes the translation hypothesis and tests whether a metric ranks a correct translation above a phenomenon-specific incorrect alternative. We instead hold a known good hypothesis fixed and intervene separately on the other two inputs: we replace its source with an incompatible source and its reference with an incompatible reference. Comparing the resulting score changes provides a paired measure of source sensitivity and reference sensitivity. This diagnostic therefore measures which input governs a hybrid metric's score, rather than whether the metric succeeds on a particular class of translation errors. It operationalizes our central distinction between \emph{source access} and \emph{source grounding}: a metric is source-grounded only if its adequacy judgment responds appropriately to the source, particularly when source and reference evidence conflict.

\section{Adequacy and Fluency}
\label{sec:adequacy-fluency}

Machine translation quality has traditionally been characterized along two broad dimensions: adequacy and fluency. Adequacy concerns the extent to which a translation preserves the meaning of its source, whereas fluency concerns whether the output is grammatical, natural, and acceptable in the target language. This distinction has a long history in human MT evaluation \citep{koehn-monz-2006-manual,snover-etal-2009-fluency} and remains central to contemporary evaluation \citep{popovic-2020-informative,shayegh-etal-2025-feeding}.

Although adequacy and fluency are related, they require different evidence. Fluency is primarily a property of the target-language output and can therefore be assessed largely from the hypothesis alone. Adequacy is relational: it concerns whether the hypothesis preserves the information expressed by the source. A translation may be perfectly fluent while omitting content, reversing a semantic relation, changing a named entity, introducing unsupported information, or resolving an ambiguity not resolved by the source. Such errors may be impossible to identify from the target sentence alone. This problem has become particularly prominent with neural and LLM-based MT, whose fluent outputs can conceal adequacy errors that become apparent only when compared with the source \citep{ustaszewski-2019-exploring}.

Reference-based evaluation approximates this judgment by comparing the hypothesis with one or more human translations. This approach underlies classical automatic metrics such as BLEU and METEOR \citep{papineni-etal-2002-bleu,banerjee-lavie-2005-meteor}. A reference provides useful evidence because it demonstrates one acceptable realization of the source meaning. However, it does not provide a complete specification of that meaning. The same source can often be translated through different lexical choices, syntactic structures, discourse organizations, or levels of explicitness. Consequently, disagreement with a reference does not necessarily indicate an adequacy error. This limitation has been recognized in early critiques of reference-overlap metrics \citep{callison-burch-etal-2006-evaluating} and in later studies of acceptable translation variation and reference construction \citep{fomicheva-bel-2016-using,freitag-etal-2020-bleu}.

More fundamentally, a reference may resolve a distinction that the source itself leaves unresolved. Consider the source sentence:

\begin{quote}
\small
\textit{Ravi told Mohan that his proposal had been rejected.}
\end{quote}

The possessive \textit{his} may refer to either Ravi or Mohan. Without additional context, the source does not determine which proposal was rejected. A human reference must nevertheless produce a particular target-language sentence. It may preserve the ambiguity, or it may select one interpretation because of translator preference, target-language convention, or grammatical necessity. If the reference selects Mohan as the antecedent, a hypothesis that instead preserves the ambiguity or selects Ravi should not automatically be considered inadequate. The source, rather than the reference, determines which interpretations are licensed.
This example illustrates that adequacy involves preserving not only what the source explicitly states, but also what it leaves unspecified. When the source contains a genuine ambiguity and the target language allows that ambiguity to be retained naturally, a faithful translation should preserve it. Resolving the ambiguity without contextual evidence introduces information that is absent from the source. Although word-sense disambiguation has long been studied as part of MT \citep{carpuat-wu-2007-improving}, disambiguation is appropriate only when the source or its context supports the selected interpretation. An evaluator should therefore distinguish justified disambiguation from an unsupported semantic commitment.

Preserving ambiguity is not always possible. Languages differ in the distinctions that their grammars require speakers to express. A source language may leave gender, number, politeness, evidentiality, or other semantic features unspecified, while the target language may require one of these features to be marked. Gender is a prominent example: English often permits gender-neutral expressions in contexts where languages with grammatical gender require gender-specific agreement. In such cases, every natural translation may commit to information not explicitly present in the source, or preserving neutrality may require substantial paraphrasing \citep{vanmassenhove-etal-2018-getting,savoldi-etal-2021-gender,dawkins-etal-2025-gender}.
Adequacy must therefore be assessed relative to both the source meaning and the expressive constraints of the target language. Ambiguity should be preserved when a natural meaning-equivalent formulation permits it. When the target language forces a choice, multiple realizations may be equally compatible with the source. A single reference records only the choice made by one translator and should not elevate that choice into a source-supported fact. Penalizing every alternative conflates faithfulness to the source with conformity to the reference.

The reverse problem is equally important. Agreement with a reference does not establish adequacy when the hypothesis and reference share an interpretation that is unsupported by, or inconsistent with, the source. A source-free evaluator can measure target-side compatibility, but it cannot determine whether their shared interpretation faithfully translates the original. Reference-free adequacy evaluation has therefore also been formulated directly as semantic compatibility between the source and hypothesis \citep{mehdad-etal-2012-match}. Under this formulation, the source is not merely a substitute for an unavailable reference. It is the evidence that defines the adequacy judgment.
This distinction remains important for learned metrics that receive the source, hypothesis, and reference together. Source access is not equivalent to source grounding. A metric may formally accept all three inputs while relying predominantly on hypothesis-reference compatibility. Challenge-set analyses show that many contemporary metrics insufficiently use the source, prefer surface-level overlap, and struggle with translation phenomena whose correct evaluation requires source-side information \citep{moghe-etal-2025-machine}. When the source and reference support different judgments, a source-grounded evaluator should treat the source as authoritative and the reference as auxiliary evidence.

Throughout this paper, we therefore use \emph{adequacy} in a strict relational sense: the degree to which a hypothesis preserves the meaning licensed by its source, including its ambiguity and underspecification where these can reasonably be preserved. Fluency may be assessed from the target text, and reference agreement may provide useful supporting evidence, but neither establishes source faithfulness. Source-free evaluation is therefore structurally incomplete as an evaluation of adequacy. The source is not optional context; it is the object against which translation adequacy must be judged.

\section{Hybrid Metrics are Biased Towards References}
Hybrid MT metrics are designed to use the source, the hypothesis, and the reference together. In principle, this should make them better suited for adequacy evaluation than purely reference-based metrics, because the source is available to check whether the hypothesis preserves the input meaning. However, the presence of the source in the metric input does not guarantee that the metric uses it as the primary evidence for adequacy. A metric may still behave as if the reference is the main standard, while using the source only weakly or inconsistently. We therefore ask whether hybrid metrics are more sensitive to the source or to the reference when scoring a known good translation.

To test this, we use a counterfactual replacement diagnostic. For each example $i$, we take the source sentence $s_i$, the good translation $g_i$, and the reference $r_i$. We then sample another example $j$ from the same language pair and construct two corrupted inputs. In the source-replaced condition, we score the same good translation and reference but replace the source with $s_j$, i.e., $M(s_j, g_i, r_i)$. In the reference-replaced condition, we keep the original source and good translation but replace the reference with $r_j$, i.e., $M(s_i, g_i, r_j)$. The original score is $M(s_i, g_i, r_i)$.
\begin{figure}[t]
    \centering
    \includegraphics[
        width=0.7\linewidth
    ]{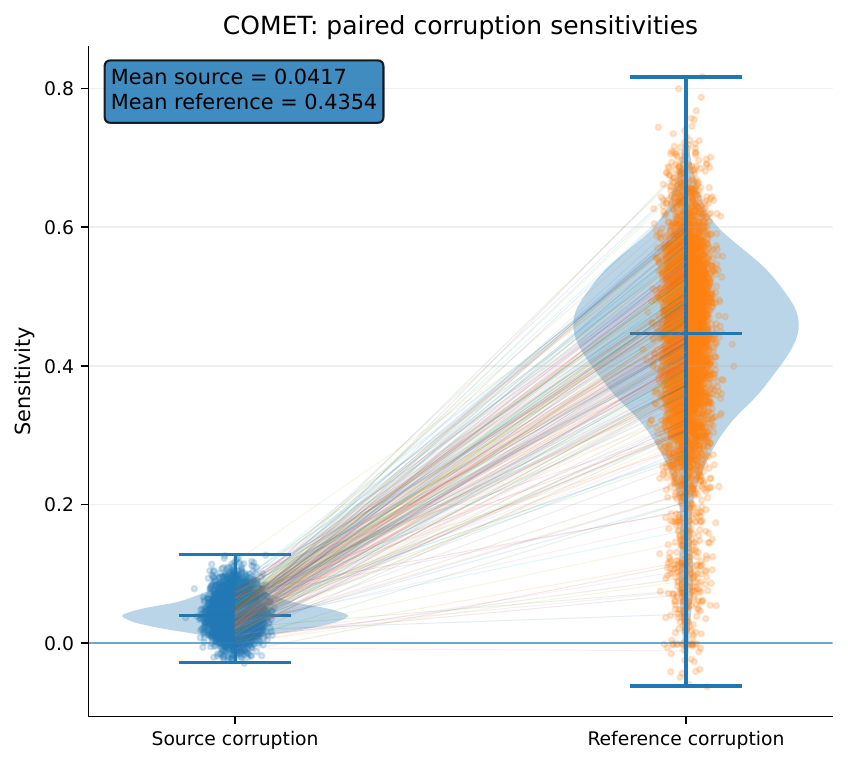}
    \caption{Paired source- and reference-corruption sensitivities for COMET.
    Positive sensitivity denotes a reduction in score after corruption. Each
    pair corresponds to the same hypothesis. Reference
    corruption produces a substantially larger score reduction for almost
    every example.}
    \label{fig:comet-paired-sensitivity}
\end{figure}

This setup isolates the relative effect of source and reference corruption while keeping the candidate translation fixed. If a hybrid metric primarily judges adequacy with respect to the source, then replacing the source should cause a large drop in quality, because the good translation $g_i$ is no longer a translation of the corrupted source $s_j$. In contrast, replacing the reference should be less damaging, because the original source $s_i$ and the good translation $g_i$ still form a valid translation pair. Conversely, if replacing the reference damages the score more than replacing the source, the metric is behaving more like a reference-sensitive metric than a truly source-grounded adequacy metric.

We quantify this behavior using the Ref/Src gap ratio. For higher-is-better metrics such as COMET and XCOMET-XXL, this ratio compares the score drop caused by reference replacement with the score drop caused by source replacement. For lower-is-better metrics such as MetricX-24-Hybrid-XL, the ratio compares the corresponding score increases. A value above 1 means that replacing the reference is more damaging than replacing the source. This is evidence that the metric is more sensitive to the reference than to the source. Ideally, the ratio should be substantially below 1, indicating that the metric is primarily grounded in the source and uses the reference as auxiliary evidence.

\subsection{Source--Reference Sensitivity of COMET}

We evaluate COMET on 6,902 retained ACES examples from 14 translation
directions. Twelve directions contain 500 examples each; \texttt{en-ja} and
\texttt{ru-en} contain 430 and 472 examples, respectively.

\paragraph{Overall sensitivity}

COMET assigns a mean score of 0.8816 to the original, internally consistent
inputs. Replacing the source reduces this score to 0.8399, producing a mean
source sensitivity of 0.0417. Replacing the reference reduces the mean score
to 0.4462, giving a mean reference sensitivity of 0.4354. Reference
replacement is therefore \(10.43\times\) as damaging as source replacement.

This asymmetry is present at the level of individual examples. Reference
sensitivity exceeds source sensitivity for 6,870 of the 6,902 examples, or
99.54\%. Only 32 examples show the opposite or no difference. The medians
closely follow the means: median source and reference sensitivities are
0.0404 and 0.4468, respectively. The aggregate result is consequently not
produced by a small number of unusually large reference effects.

\autoref{fig:comet-paired-sensitivity} makes the separation visible. Source
sensitivity is narrowly concentrated near 0.04, whereas reference sensitivity
is centered more than an order of magnitude higher. More importantly, the
within-example comparisons overwhelmingly move in the same direction:
changing the reference has the larger effect for almost every candidate.

\begin{table}[t]
    \centering
    \scriptsize
    \setlength{\tabcolsep}{3.5pt}
    \begin{tabular}{lrrrrrrr}
        \toprule
        & \multicolumn{3}{c}{Mean COMET score}
        & \multicolumn{4}{c}{Sensitivity} \\
        \cmidrule(lr){2-4}
        \cmidrule(lr){5-8}
        Language pair
        & Original
        & \shortstack{Source\\replaced}
        & \shortstack{Reference\\replaced}
        & \(\Delta_{\mathrm{src}}\)
        & \(\Delta_{\mathrm{ref}}\)
        & Ref/Src
        & \shortstack{Ref.\\dominates} \\
        \midrule
        \multicolumn{8}{c}{\emph{Into English}} \\
        \midrule
        \texttt{de-en} & 0.8898 & 0.8459 & 0.4280 & 0.0438 & 0.4618 & \(10.53\times\) & 100.0\% \\
        \texttt{es-en} & 0.8992 & 0.8527 & 0.3959 & 0.0465 & 0.5033 & \(10.83\times\) &  99.8\% \\
        \texttt{fr-en} & 0.8957 & 0.8538 & 0.4206 & 0.0418 & 0.4751 & \(11.36\times\) & 100.0\% \\
        \texttt{it-en} & 0.8846 & 0.8561 & 0.5027 & 0.0285 & 0.3819 & \(13.41\times\) &  99.8\% \\
        \texttt{ja-en} & 0.9078 & 0.8720 & 0.4542 & 0.0358 & 0.4536 & \(12.65\times\) & 100.0\% \\
        \texttt{ko-en} & 0.9102 & 0.8701 & 0.4524 & 0.0400 & 0.4577 & \(11.44\times\) &  99.8\% \\
        \texttt{ru-en} & 0.8573 & 0.8095 & 0.4381 & 0.0478 & 0.4192 & \( 8.77\times\) & 100.0\% \\
        \texttt{zh-en} & 0.8927 & 0.8534 & 0.4291 & 0.0394 & 0.4636 & \(11.78\times\) & 100.0\% \\
        \midrule 
        \multicolumn{8}{c}{\emph{Out of English}} \\
        \midrule
        \texttt{en-de} & 0.7815 & 0.7340 & 0.3833 & 0.0476 & 0.3982 & \( 8.37\times\) &  99.6\% \\
        \texttt{en-es} & 0.8685 & 0.8113 & 0.4089 & 0.0571 & 0.4595 & \( 8.04\times\) & 100.0\% \\
        \texttt{en-fr} & 0.8795 & 0.8267 & 0.3832 & 0.0528 & 0.4963 & \( 9.40\times\) & 100.0\% \\
        \texttt{en-ja} & 0.9031 & 0.8685 & 0.5576 & 0.0345 & 0.3455 & \(10.01\times\) &  99.1\% \\
        \texttt{en-ko} & 0.8909 & 0.8469 & 0.4973 & 0.0440 & 0.3935 & \( 8.95\times\) &  99.4\% \\
        \texttt{en-ru} & 0.8839 & 0.8597 & 0.5111 & 0.0242 & 0.3728 & \(15.43\times\) &  96.0\% \\
        \bottomrule
    \end{tabular}
    \caption{COMET scores and corruption sensitivities by language pair.
    Original is the mean score for the consistent source, good translation,
    and reference. Source replaced and Reference replaced report the mean
    scores after replacing the corresponding input. Ref/Src is the ratio of
    mean reference sensitivity to mean source sensitivity. Ref.\ dominates
    is the percentage of examples for which reference sensitivity exceeds
    source sensitivity.}
    \label{tab:comet-langpair-sensitivity}
\end{table}
\paragraph{Consistency across translation directions}

The aggregate pattern holds both into and out of English. For the 3,972
into-English examples, mean source and reference sensitivities are 0.0404 and
0.4523, producing a Ref/Src ratio of \(11.19\times\). Reference sensitivity
is larger for 99.92\% of these examples. For the 2,930 out-of-English
examples, the corresponding sensitivities are 0.0436 and 0.4125, giving a
ratio of \(9.47\times\) and a reference-dominance rate of 99.01\%. The
difference between the two groups is therefore one of magnitude, not a
reversal of behavior.

\autoref{tab:comet-langpair-sensitivity} provides the finer language-pair
breakdown. Reference corruption has the larger mean effect in every examined
direction. The Ref/Src ratio never falls below \(8\times\), and the
example-level reference-dominance rate never falls below 96\%.

The ratio and dominance columns provide complementary information. The
smallest ratio occurs for \texttt{en-es}, yet reference corruption remains
\(8.04\times\) as damaging and dominates on every example. Conversely,
\texttt{en-ru} has the largest ratio, \(15.43\times\), but the lowest
dominance rate, 96.0\%. Its average asymmetry is exceptionally large but less
uniform across examples. Thus, neither the magnitude nor the consistency of
the overall result depends on a particular language pair.

\paragraph{Source sensitivity versus source grounding}

The result does not imply that COMET ignores its source input. Source
replacement lowers the score for 98.17\% of examples, indicating that the
source normally affects the metric. Reference replacement lowers it for
99.59\% of examples. The distinction lies in the size of these effects:
source corruption typically causes a small reduction, while reference
corruption causes a reduction more than ten times larger.

COMET is therefore source-sensitive in an absolute sense but
reference-dominated in a relative sense. Its score responds to an incompatible
source, yet responds far more strongly to an incompatible reference, including
when the unchanged source and candidate remain a valid translation pair. This
gap between using the source and being grounded in it is precisely the concern
that input access alone cannot resolve.
\begin{figure}
    \centering
    \includegraphics[
        width=0.7\linewidth
    ]{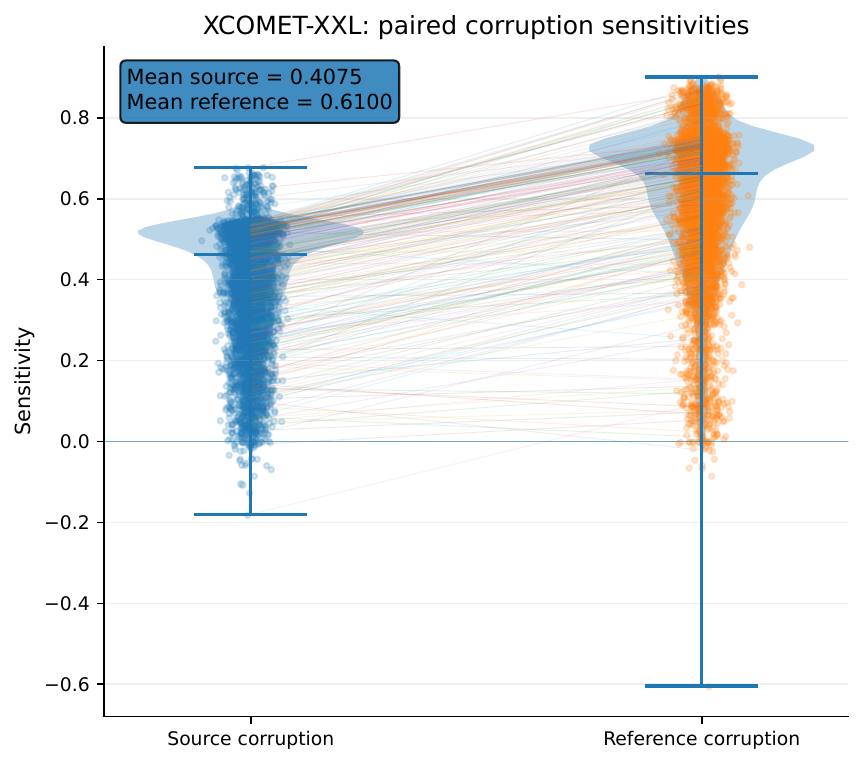}
    \caption{Paired source- and reference-corruption sensitivities for
    XCOMET-XXL. Positive sensitivity denotes a reduction in score after
    corruption. Each pair corresponds to the same hypothesis.
    Reference corruption produces a larger score reduction for the great
    majority of examples.}
    \label{fig:xcomet-paired-sensitivity}
\end{figure}
\subsection{Source--Reference Sensitivity of XCOMET-XXL}
\begin{table}
    \centering
    \scriptsize
    \setlength{\tabcolsep}{3.5pt}
    \begin{tabular}{lrrrrrrr}
        \toprule
        & \multicolumn{3}{c}{Mean XCOMET-XXL score}
        & \multicolumn{4}{c}{Sensitivity} \\
        \cmidrule(lr){2-4}
        \cmidrule(lr){5-8}
        Language pair
        & Original
        & \shortstack{Source\\replaced}
        & \shortstack{Reference\\replaced}
        & \(\Delta_{\mathrm{src}}\)
        & \(\Delta_{\mathrm{ref}}\)
        & Ref/Src
        & \shortstack{Ref.\\dominates} \\
        \midrule
        \multicolumn{8}{c}{\emph{Into English}} \\
        \midrule
        \texttt{de-en} & 0.8660 & 0.4468 & 0.2364 & 0.4191 & 0.6295 & \(1.50\times\) &  97.4\% \\
        \texttt{es-en} & 0.9115 & 0.4398 & 0.2305 & 0.4717 & 0.6810 & \(1.44\times\) &  99.0\% \\
        \texttt{fr-en} & 0.8130 & 0.4373 & 0.2205 & 0.3758 & 0.5925 & \(1.58\times\) &  96.6\% \\
        \texttt{it-en} & 0.7035 & 0.4443 & 0.2197 & 0.2592 & 0.4837 & \(1.87\times\) &  98.2\% \\
        \texttt{ja-en} & 0.8784 & 0.4405 & 0.2169 & 0.4379 & 0.6615 & \(1.51\times\) &  99.6\% \\
        \texttt{ko-en} & 0.8326 & 0.4092 & 0.2204 & 0.4234 & 0.6122 & \(1.45\times\) &  90.4\% \\
        \texttt{ru-en} & 0.9036 & 0.4527 & 0.2501 & 0.4510 & 0.6535 & \(1.45\times\) &  99.6\% \\
        \texttt{zh-en} & 0.9276 & 0.4613 & 0.2386 & 0.4663 & 0.6890 & \(1.48\times\) & 100.0\% \\
        \midrule 
        \multicolumn{8}{c}{\emph{Out of English}} \\
        \midrule
        \texttt{en-de} & 0.8467 & 0.4275 & 0.2410 & 0.4192 & 0.6057 & \(1.44\times\) & 93.8\% \\
        \texttt{en-es} & 0.9200 & 0.4359 & 0.2083 & 0.4841 & 0.7117 & \(1.47\times\) & 99.2\% \\
        \texttt{en-fr} & 0.8704 & 0.4139 & 0.2234 & 0.4565 & 0.6470 & \(1.42\times\) & 98.4\% \\
        \texttt{en-ja} & 0.8051 & 0.3970 & 0.2209 & 0.4082 & 0.5843 & \(1.43\times\) & 98.1\% \\
        \texttt{en-ko} & 0.8414 & 0.4171 & 0.2198 & 0.4242 & 0.6215 & \(1.47\times\) & 97.6\% \\
        \texttt{en-ru} & 0.7164 & 0.5053 & 0.3506 & 0.2111 & 0.3657 & \(1.73\times\) & 79.0\% \\
        \bottomrule
    \end{tabular}
    \caption{XCOMET-XXL scores and corruption sensitivities by language pair.
    Original is the mean score for the consistent source, good translation,
    and reference. Source replaced and Reference replaced report the mean
    scores after replacing the corresponding input. Ref/Src is the ratio of
    mean reference sensitivity to mean source sensitivity. Ref.\ dominates
    is the percentage of examples for which reference sensitivity exceeds
    source sensitivity. Each direction contains 500 examples except
    \texttt{en-ja}, with 430, and \texttt{ru-en}, with 472.}
    \label{tab:xcomet-langpair-sensitivity}
\end{table}
We apply the same diagnostic to XCOMET-XXL over the same 6,902 examples and
14 translation directions.
XCOMET-XXL assigns a mean score of 0.8456 to the original inputs. Source
replacement reduces the mean score to 0.4381, corresponding to a source
sensitivity of 0.4075. Reference replacement reduces it further to 0.2356,
corresponding to a reference sensitivity of 0.6100. Thus, unlike COMET, for
which source corruption causes only a small score change, XCOMET-XXL responds
strongly to both inputs. Nevertheless, reference replacement remains
\(1.50\times\) as damaging as source replacement.

The difference is highly consistent across examples. Reference sensitivity
exceeds source sensitivity for 6,638 of the 6,902 examples, giving a
reference-dominance rate of 96.18\%. The median source and reference
sensitivities are 0.4623 and 0.6621, respectively. The agreement between the
mean and median differences indicates that the aggregate asymmetry is not
produced by a small number of extreme cases.

\paragraph{Paired sensitivity distribution}

\autoref{fig:xcomet-paired-sensitivity} shows that both corruption conditions
produce substantial score reductions, but the reference-sensitivity
distribution is consistently shifted upward. The distributions overlap more
than they do for COMET, which is expected from the smaller Ref/Src ratio.
However, the paired comparisons reveal that reference corruption still has
the larger effect for the great majority of individual candidates.

\paragraph{Consistency across translation directions}

Reference dominance remains stable after separating translations into and out
of English. For the 3,972 into-English examples, source and reference
sensitivities are 0.4128 and 0.6252, respectively. This gives a Ref/Src ratio
of \(1.51\times\), with reference sensitivity exceeding source sensitivity
for 97.58\% of examples. For the 2,930 out-of-English examples, the
corresponding sensitivities are 0.4004 and 0.5894, producing a ratio of
\(1.47\times\) and a reference-dominance rate of 94.27\%. The aggregate
finding is therefore not attributable to the larger number of into-English
examples.

The language-pair results in
\autoref{tab:xcomet-langpair-sensitivity} show the same qualitative pattern.
Mean reference sensitivity exceeds mean source sensitivity in all 14
directions. The Ref/Src ratios range from \(1.42\times\) for
\texttt{en-fr} to \(1.87\times\) for \texttt{it-en}.

The example-level dominance rates range from 79.0\% to 100.0\%. The lowest
rate occurs for \texttt{en-ru}, despite this direction having the
second-largest Ref/Src ratio, \(1.73\times\). This indicates a comparatively
heterogeneous distribution: reference corruption has a substantially larger
mean effect, but a larger minority of individual examples show equal or
greater source sensitivity. In every other direction, reference corruption
has the larger effect for at least 90.4\% of examples.

\paragraph{Interpretation}

XCOMET-XXL makes considerably greater use of the source than COMET under this
diagnostic. Source replacement lowers its score for 99.38\% of examples and
causes a large mean reduction of 0.4075. The source is therefore neither
ignored nor merely weakly active. However, reference replacement lowers the
score for 99.59\% of examples, causes a larger mean reduction of 0.6100, and
dominates the paired comparison in 96.18\% of cases.

XCOMET-XXL is consequently more source-responsive than COMET, but it remains
systematically reference-dominated. Its stronger source sensitivity narrows
the magnitude of the asymmetry without changing its direction or
example-level consistency. Providing more explicit source-based modeling
therefore improves source use, but does not by itself make the source the
primary basis of the metric's adequacy judgment.

\subsection{Source--Reference Sensitivity of MetricX-24-Hybrid-XL}

We next apply the diagnostic to MetricX-24-Hybrid-XL over the same 6,902
examples and 14 translation directions. MetricX is a lower-is-better metric,
so sensitivity is defined as the increase in score caused by corruption.
Positive values therefore indicate that the corrupted input receives a worse
score.

\paragraph{Overall sensitivity}

The mean MetricX score is 3.2295 for the original consistent inputs. Source
replacement increases it to 12.4672, corresponding to a mean source
sensitivity of 9.2377. Reference replacement increases it further to 15.5908,
corresponding to a mean reference sensitivity of 12.3613. The aggregate
Ref/Src ratio is therefore \(1.34\times\).

Reference sensitivity exceeds source sensitivity for 5,170 of the 6,902
examples, giving a reference-dominance rate of 74.91\%. The median source and
reference sensitivities are 9.2188 and 12.7813, respectively. Thus, the
aggregate difference is visible in both the mean and the median, although it
is substantially less uniform than for COMET or XCOMET-XXL.

\paragraph{Paired sensitivity distribution}

\begin{figure}[t]
    \centering
    \includegraphics[
        width=0.7\linewidth
    ]{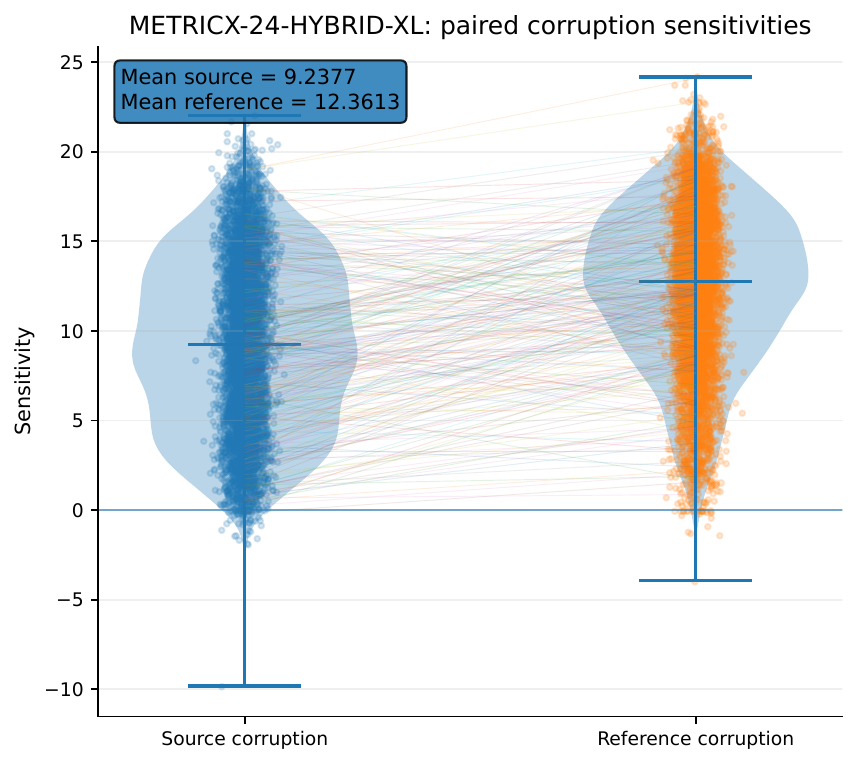}
    \caption{Paired source- and reference-corruption sensitivities for
    MetricX-24-Hybrid-XL. Positive sensitivity denotes an increase in score
    after corruption because lower MetricX scores indicate better translation
    quality.
    Reference corruption has the larger effect overall, but the two
    sensitivity distributions overlap substantially.}
    \label{fig:metricx-paired-sensitivity}
\end{figure}

The paired distributions in
\autoref{fig:metricx-paired-sensitivity} show substantial sensitivity to both
inputs. Reference sensitivity is shifted upward overall, but the two
distributions overlap considerably and many paired lines cross. This agrees
with the example-level result: reference corruption is usually more damaging,
but source corruption has the larger effect for approximately one-quarter of
the examples.

\paragraph{Strong directional asymmetry}

The aggregate result conceals a sharp difference between translation
directions. For the 3,972 into-English examples, source and reference
sensitivities are 8.0752 and 13.5536, respectively. Reference corruption is
therefore \(1.68\times\) as damaging and has the larger effect for 95.59\% of
examples.

For the 2,930 out-of-English examples, source and reference sensitivities are
10.8136 and 10.7451. The resulting Ref/Src ratio is \(0.99\times\), and
reference sensitivity is larger for only 46.86\% of examples. MetricX is
therefore strongly reference-dominated into English but approximately
balanced, with a slight source-dominant tendency, out of English.

The language-pair results in
\autoref{tab:metricx-langpair-sensitivity} make this asymmetry explicit. All
eight into-English directions have Ref/Src ratios above 1, with
reference-dominance rates between 91.0\% and 99.4\%. In contrast, only two of
the six out-of-English directions have ratios above 1.

\begin{table}[t]
    \centering
    \scriptsize
    \setlength{\tabcolsep}{3.5pt}
    \begin{tabular}{lrrrrrrr}
        \toprule
        & \multicolumn{3}{c}{Mean MetricX score}
        & \multicolumn{4}{c}{Sensitivity} \\
        \cmidrule(lr){2-4}
        \cmidrule(lr){5-8}
        Language pair
        & Original
        & \shortstack{Source\\replaced}
        & \shortstack{Reference\\replaced}
        & \(\Delta_{\mathrm{src}}\)
        & \(\Delta_{\mathrm{ref}}\)
        & Ref/Src
        & \shortstack{Ref.\\dominates} \\
        \midrule
        \multicolumn{8}{c}{\emph{Into English}} \\
        \midrule
        \texttt{de-en} & 3.2048 &  9.7065 & 16.7304 &  6.5017 & 13.5257 & \(2.08\times\) & 99.4\% \\
        \texttt{es-en} & 2.9215 & 13.4021 & 16.8966 & 10.4805 & 13.9750 & \(1.33\times\) & 91.0\% \\
        \texttt{fr-en} & 3.5234 & 12.1798 & 16.8541 &  8.6564 & 13.3307 & \(1.54\times\) & 94.0\% \\
        \texttt{it-en} & 3.8907 &  8.3780 & 15.6120 &  4.4873 & 11.7213 & \(2.61\times\) & 98.6\% \\
        \texttt{ja-en} & 2.8538 & 11.4693 & 16.9521 &  8.6155 & 14.0983 & \(1.64\times\) & 96.4\% \\
        \texttt{ko-en} & 2.8187 & 12.1403 & 16.8473 &  9.3216 & 14.0287 & \(1.50\times\) & 94.8\% \\
        \texttt{ru-en} & 3.3644 & 12.1531 & 16.5096 &  8.7888 & 13.1452 & \(1.50\times\) & 91.9\% \\
        \texttt{zh-en} & 1.7822 &  9.5716 & 16.3631 &  7.7894 & 14.5809 & \(1.87\times\) & 98.4\% \\
        \midrule
        \multicolumn{8}{c}{\emph{Out of English}} \\
        \midrule
        \texttt{en-de} & 3.9605 & 11.9535 & 10.3840 &  7.9930 &  6.4235 & \(0.80\times\) & 30.2\% \\
        \texttt{en-es} & 2.9008 & 17.1997 & 16.2528 & 14.2989 & 13.3520 & \(0.93\times\) & 36.4\% \\
        \texttt{en-fr} & 2.8615 & 15.2658 & 16.8942 & 12.4043 & 14.0327 & \(1.13\times\) & 72.2\% \\
        \texttt{en-ja} & 3.8919 & 16.1983 & 14.1156 & 12.3064 & 10.2237 & \(0.83\times\) & 21.6\% \\
        \texttt{en-ko} & 3.6667 & 16.6214 & 15.2865 & 12.9547 & 11.6198 & \(0.90\times\) & 28.2\% \\
        \texttt{en-ru} & 3.6721 &  8.8057 & 12.4180 &  5.1336 &  8.7459 & \(1.70\times\) & 89.0\% \\
        \bottomrule
    \end{tabular}
    \caption{MetricX-24-Hybrid-XL scores and corruption sensitivities by language pair. Lower MetricX scores are better. Original is the mean score for the consistent source, good translation, and reference. Source replaced and Reference replaced report the mean scores after replacing the corresponding input. Ref/Src is the ratio of mean reference sensitivity to mean source sensitivity. Ref. dominates is the percentage of examples for which reference sensitivity exceeds source sensitivity.}
    \label{tab:metricx-langpair-sensitivity}
\end{table}

The contrast is particularly clear for \texttt{en-ja},
\texttt{en-de}, and \texttt{en-ko}. Their Ref/Src ratios are
\(0.83\times\), \(0.80\times\), and \(0.90\times\), while their
reference-dominance rates are only 21.6\%, 30.2\%, and 28.2\%. Source
corruption is therefore more consequential for most examples in these
directions. Conversely, \texttt{en-ru} remains strongly reference-dominated,
with a ratio of \(1.70\times\) and a dominance rate of 89.0\%.
Out-of-English behavior is thus heterogeneous rather than uniformly
source-dominated.

\paragraph{Interpretation}

MetricX is highly responsive to both inputs. Source replacement worsens the
score for 99.25\% of examples, while reference replacement worsens it for
99.62\%. Its aggregate reference dominance therefore does not arise from
ignoring the source. Instead, it arises primarily from the strong reference
effects observed when English is the target language.

Unlike the patterns observed for COMET and XCOMET-XXL, MetricX cannot be
adequately characterized by a single pooled ratio. The overall
\(1.34\times\) ratio combines two qualitatively different regimes:
substantial reference dominance into English and approximate source-reference
balance out of English. This direction dependence qualifies the broader
claim of reference bias while also showing that MetricX's reliance on the
source is not uniform across translation settings.

\subsection{Overall Observations}

The diagnostic reveals three distinct patterns of source--reference use. COMET shows the clearest reference dependence: source corruption causes only a small penalty relative to reference corruption, and this behavior is nearly universal across examples and language pairs. XCOMET-XXL reacts much more strongly to source corruption than COMET, suggesting that its explicit error-detection training increases its use of source information. Nevertheless, reference corruption remains more damaging for the large majority of examples, so the reference still acts as the stronger signal.

MetricX exhibits the most direction-dependent behavior. Its aggregate result conceals a sharp contrast: it is strongly reference-dominated into English, but source and reference corruption have similar average effects out of English. The language-pair results further show that this out-of-English balance is not uniform, since individual directions range from source-dominated to strongly reference-dominated. MetricX therefore cannot be characterized by a single, direction-independent source--reference preference.

The Ref/Src ratio and example-level dominance rate capture complementary aspects of these patterns. The ratio measures the relative magnitude of the two corruption effects, whereas the dominance rate measures how consistently their ordering holds across examples. XCOMET-XXL, for example, has a much smaller ratio than COMET but still exhibits reference dominance on most examples. MetricX demonstrates why both quantities are necessary, since similar average effects can coexist with substantial example-level and language-pair variation.

One possible explanation for MetricX's directional asymmetry is greater reliance on whichever input component is expressed in English. This possibility is compatible with its fine-tuning setup, which includes several MQM directions with English as the source \citep{juraska-etal-2024-metricx}. However, our experiment does not establish this explanation because translation direction confounds input role with language identity. Non-English-to-non-English directions would be needed to determine whether the pattern reflects English familiarity, training-data composition, or some other model property.

Overall, providing a metric with the source does not ensure that the source becomes its primary evidence for adequacy. COMET and XCOMET-XXL remain systematically more dependent on the reference, while MetricX's dependence changes with translation direction. These results do not imply that reference information is unhelpful, but they show that hybrid input access alone is insufficient evidence of genuinely source-grounded evaluation.

\section{LLMs as Judges and the Source-Free Evaluation Problem}
\label{sec:llm-judge}

Large language models have increasingly been used as direct judges of machine translation quality. GEMBA showed that sufficiently capable GPT models can produce competitive system-level evaluations through zero-shot prompting, both with and without a human reference \citep{kocmi-federmann-2023-large}. Subsequent methods moved from predicting a single score toward identifying and classifying translation errors. AutoMQM prompts an LLM to locate errors and assign MQM categories and severities \citep{fernandes-etal-2023-devil}; GEMBA-MQM applies a similar error-annotation protocol using GPT-4 in a reference-free setting \citep{kocmi-federmann-2023-gemba}; and EAPrompt combines error analysis with structured reasoning to improve segment-level evaluation \citep{lu-etal-2024-error}. More recently, M-MAD decomposes evaluation into separate accuracy, fluency, style, and terminology dimensions and uses multi-agent debate to consolidate the resulting judgments \citep{feng-etal-2025-mad}. Together, these approaches suggest that LLMs can act not only as scalar metrics, but also as flexible evaluators capable of explaining their decisions.

The flexibility of prompting, however, does not resolve the distinction between reference-free and source-free evaluation. A reference-free LLM judge receives the source and hypothesis but no reference. A source-free judge receives the hypothesis and reference but not the source. The former retains the information necessary to assess adequacy, although the judge may fail to use it effectively. The latter removes the only direct evidence of what was to be translated. A third setting supplies the source, hypothesis, and reference together, but this setting is source-grounded only if the source governs the adequacy judgment when the source and reference provide conflicting evidence.

Controlled experiments indicate that current LLM judges do not reliably satisfy this condition. \citet{huang-etal-2024-lost} compare four prompting configurations: translation only (\textsc{T}), source and translation (\textsc{S-T}), reference and translation (\textsc{R-T}), and source, reference, and translation (\textsc{S-R-T}). Across multiple models, the reference generally contributes more to agreement with human judgments than the source. For GPT-3.5-Turbo, the \textsc{R-T} configuration obtains a system-level accuracy of 0.891, whereas adding the source in the \textsc{S-R-T} configuration reduces it to 0.876. Segment-level Kendall correlations also decrease. Similar behavior is observed in several open models, in fine-grained error detection, and after task-specific fine-tuning. Their Shapley-value analysis likewise finds that the reference generally makes a substantially larger contribution than the source.

Importantly, these results do not show that the source is unnecessary for MT evaluation. They show that current LLMs are better at exploiting monolingual target-side comparison than at reasoning reliably across languages. Removing the source because its inclusion lowers correlation would optimize the evaluator around its weakness. It would transform a failure to perform source-grounded adequacy assessment into a justification for avoiding adequacy assessment altogether. Higher agreement with human scores in the \textsc{R-T} setting establishes that reference comparison is an effective predictor under the evaluated distribution; it does not establish that the resulting judgment is faithful to the source.

This distinction becomes particularly important when the source and reference disagree or when the reference is insufficient to determine the correct translation. A judge can obtain high average correlation by exploiting reference similarity while failing on omissions, additions, ambiguity, negation, named entities, or other phenomena for which the source is decisive. Using source-dependent challenge sets, \citet{moghe-etal-2025-machine} find that LLM-based evaluators are unreliable across translation phenomena and that evaluators frequently fail when the better hypothesis can only be identified from the source. Consequently, aggregate correlation and source grounding measure different properties. The first asks whether a judge predicts available human scores. The second asks whether its adequacy decision is governed by the evidence that defines translation correctness.

\subsection{Reference-Free LLM Judges}
\label{sec:reference-free-llm-judges}

Some LLM evaluation protocols avoid reference dominance by construction. GEMBA-MQM evaluates a translation from its source and hypothesis and asks GPT-4 to identify error spans and their MQM categories and severities \citep{kocmi-federmann-2023-gemba}. AutoMQM can similarly operate without a reference, prompting the judge to identify additions, omissions, mistranslations, and other errors through direct comparison with the source \citep{fernandes-etal-2023-devil}. M-MAD is also formulated as a reference-free evaluator and explicitly defines accuracy errors according to whether the target content reflects the propositional content of the source \citep{feng-etal-2025-mad}. These protocols are more consistent with the definition of translation adequacy because they preserve the source--hypothesis relationship instead of substituting hypothesis--reference similarity for it.

Reference-free design nevertheless does not by itself guarantee source grounding. An LLM may receive the source while relying primarily on target-language fluency, general plausibility, or memorized translation patterns. Indeed, the critical-error experiments of \citet{huang-etal-2024-lost} show that even strong LLMs can overlook information in the source when detecting conspicuous translation errors. Reference-free LLM judges should therefore be validated not only through correlation with human scores, but also through behavioral tests that establish whether their judgments respond to source-side meaning.

The same requirement applies more strongly to hybrid LLM judges. A prompt containing the source, hypothesis, and reference should not be assumed to provide a source-grounded evaluation merely because all three fields are present. When a reference is available, its target-language similarity provides an easier signal than cross-lingual comparison and may dominate the judgment. We therefore treat source grounding as an empirical property rather than a prompt-design label. An LLM judge is source-grounded only when its adequacy judgment changes appropriately with the source and treats the reference as auxiliary evidence when the two conflict.

\section{Recommendations}

The community should treat quality estimation as a primary approach to adequacy evaluation, rather than as a substitute used only when references are unavailable. Research should focus directly on whether a hypothesis preserves the source meaning. This requires source-grounded training data, ideally with bilingual adequacy judgments collected without showing the reference, and benchmarks containing omissions, additions, ambiguity, and valid translations that differ from the reference.

References can still provide useful auxiliary evidence, but hybrid metrics should be designed so that they cannot easily override the source. Possible approaches include separately modeling source--hypothesis adequacy and hypothesis--reference compatibility, randomly removing references during training, using multiple valid references, and training on counterfactual source and reference replacements. LLM judges could similarly assess source--hypothesis adequacy before receiving the reference. These approaches may encourage source grounding, but they are not guaranteed to work and must be validated through behavioral tests.

Metric evaluation should therefore measure source grounding separately from correlation with human scores. In addition to conventional correlation, developers should report performance on source-dependent challenge sets, source and reference ablations, and cases in which the reference is incomplete or misleading. Results should be separated by translation direction and should include non-English-to-non-English pairs, since pooled results can conceal language-specific behavior. At the same time, source sensitivity should not be maximized mechanically, because reacting to source changes does not necessarily imply understanding them.

Finally, MT papers should not present source-free scores as complete measures of translation quality. Reference-based metrics remain useful for measuring target-side agreement, but claims about adequacy should include a source-grounded evaluator whose use of the source has been empirically demonstrated. Shared tasks and leaderboards should likewise report source grounding alongside human-score correlation. The goal is not to eliminate references, but to ensure that the source remains the primary authority when translation adequacy is evaluated.

\section{Conclusion}

MT adequacy is defined by whether a hypothesis preserves the meaning of its source. Source-free evaluation can provide useful evidence about fluency and agreement with a reference, but it cannot independently establish this relationship. The same concern applies to hybrid metrics: receiving the source does not guarantee that the source governs the judgment. Our counterfactual diagnostic shows that COMET and XCOMET-XXL are systematically more sensitive to reference corruption than to source corruption, while MetricX exhibits substantial direction-dependent behavior. Existing findings on LLM judges reveal a similar risk of relying on the reference or target-side plausibility instead of source meaning. Our argument is, while references remain valuable as auxiliary evidence, but they should not override a valid source--hypothesis relationship. The community should therefore treat source grounding as the primary criterion for measuring adequacy.
\section*{Acknowledgments}
The authors used ChatGPT \cite{openai2024gpt4technicalreport} as a coding assistant for debugging and extending the core implementation to cover all language pairs considered in this study. ChatGPT was also used during brainstorming to explore clearer and more engaging ways of presenting the paper's arguments. All generated suggestions and code modifications were critically reviewed and revised by the authors.
Baban Gain gratefully acknowledges the Prime Minister's Research Fellowship (PMRF) Scheme for providing financial support and enabling this research.
\bibliography{custom,anthology-1,anthology-2}
\bibliographystyle{plainnat}

\appendix
\newpage

\end{document}